\documentclass[11pt,a4paper]{article}
\usepackage[hyperref]{acl2018}
\usepackage{times}
\usepackage{latexsym}
\usepackage{bm}
\usepackage{enumitem}
\usepackage{graphicx}
\usepackage{amsmath}
\usepackage{amsfonts}
\usepackage{booktabs}
\usepackage{url}

\usepackage{xcolor}

\aclfinalcopy % Uncomment this line for the final submission
\def\aclpaperid{1465} %  Enter the acl Paper ID here

\title{Enhancing Drug-Drug Interaction Extraction from Texts by \\Molecular Structure Information}

\author{Masaki Asada{\rm{,}} Makoto Miwa \and Yutaka Sasaki \\
  Computational Intelligence Laboratory \\
  Toyota Technological Institute \\
  2-12-1 Hisakata, Tempaku-ku, Nagoya, Aichi, 468-8511, Japan \\
  {\tt \{sd17402, makoto-miwa, yutaka.sasaki\}@toyota-ti.ac.jp} \\}

\date{}

\begin{document}
\maketitle
\begin{abstract}
We propose a novel neural method to extract drug-drug interactions (DDIs) from texts using external drug molecular structure information. We encode textual drug pairs with convolutional neural networks and their molecular pairs with graph convolutional networks (GCNs), and then we concatenate the outputs of these two networks. In the experiments, we show that GCNs can predict DDIs from the molecular structures of drugs in high accuracy and the molecular information can enhance text-based DDI extraction by 2.39 percent points in the F-score on the DDIExtraction 2013 shared task data set. 
\end{abstract}

\section{Introduction}

When drugs are concomitantly administered to a patient, the effects of the drugs may be enhanced or weakened, which may also cause side effects. These kinds of interactions are called Drug-Drug Interactions (DDIs). Several drug databases have been maintained to summarize drug and DDI information such as DrugBank~\cite{drugbank}, Therapeutic Target database~\cite{yang2016therapeutic}, 
and PharmGKB~\cite{thorn2013pharmgkb}. 
%Many newly discovered or rarely reported interactions are, however, not fully covered in these databases, and 
%they are still buried in biomedical articles. 
%
Automatic DDI extraction from texts is expected to support the maintenance of databases with high coverage and quick update to help medical experts. % deepen their understanding of DDIs and find new target DDIs.
Deep neural network-based methods have recently drawn a considerable attention~\cite{liu2016drug,sahu2017drug,Zheng2017,lim2018drug} since they show state-of-the-art performance without manual feature engineering. 

In parallel to the progress in DDI extraction from texts, Graph Convolutional Networks (GCNs) have been proposed and applied to estimate physical and chemical properties of molecular graphs such as solubility and toxicity~\cite{NIPS2015_5954,li2015gated,gilmer2017neural}. 

\begin{figure*}[t]
  \centering
  \includegraphics[width=.8\linewidth]{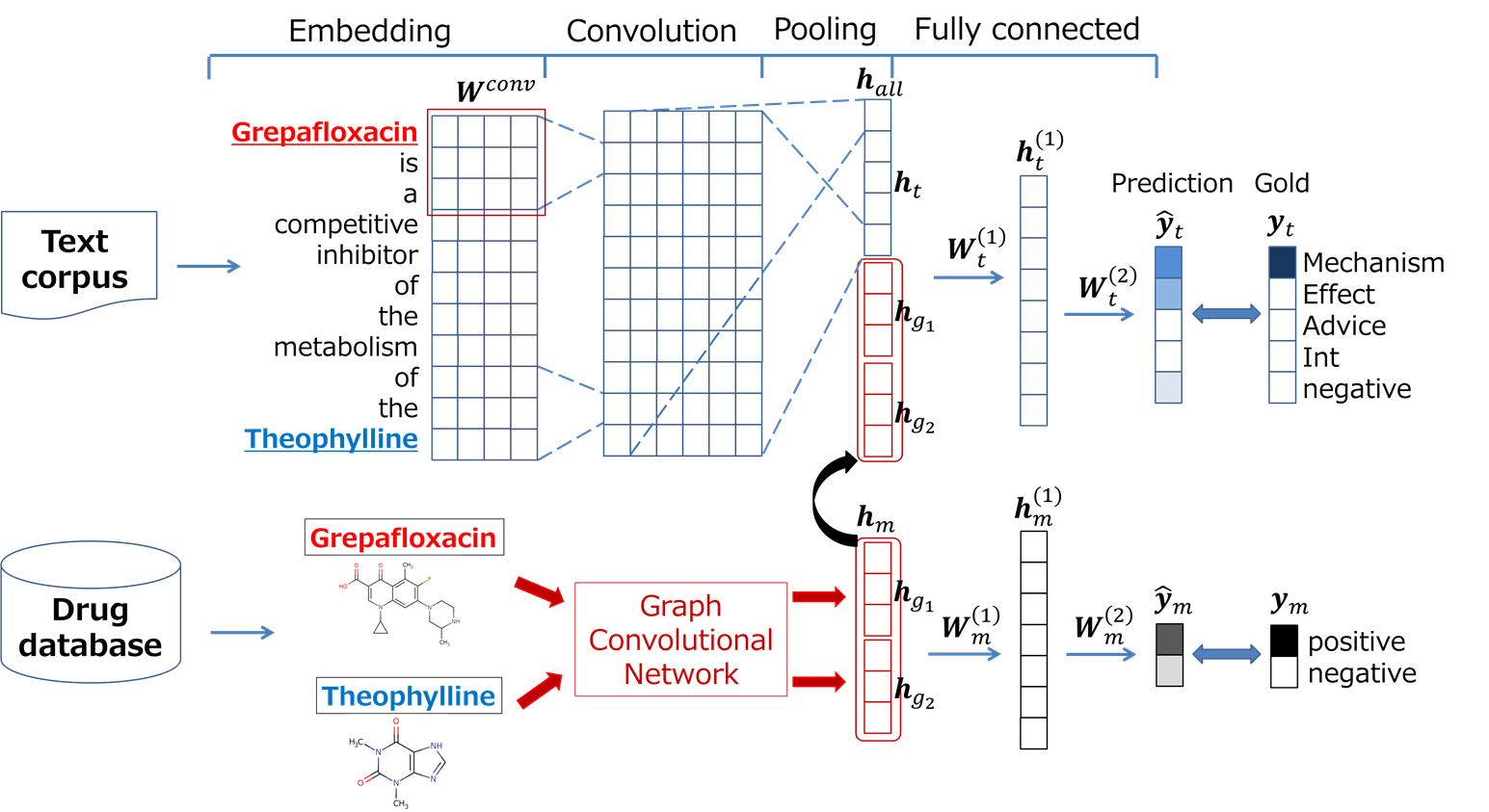}
  \caption{Overview of the proposed model}%. We omit the filters and windows of CNNs for simplicity.}
  \label{figure:model}
\end{figure*}

In this study, we propose a novel method to utilize both textual and molecular information for DDI extraction from texts. We illustrate the overview of the proposed model in Figure~\ref{figure:model}. 
We obtain the representations of drug pairs in molecular graph structures using GCNs and concatenate the representations with the representations of the textual mention pairs obtained by convolutional neural networks (CNNs). 
We trained the molecule-based model using interacting pairs mentioned in the DrugBank database and then trained the entire model using the labeled pairs in the text data set of the DDIExtraction 2013 shared task (SemEval-2013 Task 9)~\cite{segura2013semeval}.   
In the experiment, we show GCNs can predict DDIs from molecular graphs in a high accuracy. 
We also show molecular information can enhance the performance of DDI extraction from texts in 2.39 percent points in F-score. %The resulted model shows the state-of-the-art performance on the data set. 

%\todo{clarify that we firstly apply GCNs to pairwise drug interactions}
%Weakness argument 4: The authors make a claim of a contribution around the application of graph convolutional networks to pairwise drug interactions; however it is not clear that this is a novel contribution given the prior work that is mentioned in the paper.
%The graph convolutional networks have been applied only to single molecules as far as we know. 
The contribution of this paper is three-fold: 
\begin{itemize}[nolistsep]
\item{We propose a novel neural method to extract DDIs from texts with the related molecular structure information.} 
\item{We apply GCNs to pairwise drug molecules for the first time and show GCNs can predict DDIs between drug molecular structures in a high accuracy.} 
\item{We show the molecular information is useful in extracting DDIs from texts.}
\end{itemize}

\section{Methods}

\subsection{Text-based DDI Extraction}

Our model for extracting DDIs from texts is based on the CNN model by \newcite{zeng2014relation}.
When an input sentence $S=(w_1, w_2, \cdots , w_N)$ is given, 
We prepare word embedding $\bm{w}^w_i$ of $w_i$ and word position embeddings $\bm{w}^p_{i, 1}$ and $\bm {w}^p_{i,2}$ that correspond to the relative positions from the first and second target entities, respectively. 
We concatenate these embeddings as in Equation~(\ref{eq:10}), and we use the resulting vector as the input to the subsequent convolution layer:
\begin{eqnarray}
\bm{w}_i=[\bm{w}^w_i; \bm{w}^p_{i,1}; \bm{w}^p_{i,2}],
\label{eq:10}
\end{eqnarray}
where $[;]$ denotes the concatenation.
We calculate the expression for each filter $j$ with the window size $k_l$.
\begin{eqnarray}
\bm{z}_{i,l}&=&[\bm{w}_{i-(k_l-1)/2},\cdots,\bm{w}_{i-(k_l+1)/2}],\\
m_{i,j,l}&=&\mathrm{relu}(\bm{W}^{conv}_j \odot \bm{z}_{i,l}+b^{conv}), \\
m_{j,l} &=&\max_i m_{i,j,l},
\end{eqnarray}
where $L$ is the number of windows, $\bm{W}^{conv}_j$ and $b^{conv}$ are the weight and bias of CNN, and $\max$ indicates max pooling~\cite{maxpooling}.

We convert the output of the convolution layer into a fixed-size vector that represents a textual pair as follows:
\begin{eqnarray}
\bm{m}_l&=&[m_{1,l}, \cdots, m_{J,l}],\\
\bm{h}_{t}&=&[\bm{m}_{1}; \ldots; \bm{m}_{L}],
\end{eqnarray}
where $J$ is the number of filters. 

We get a prediction $\hat{\bm{y}_t}$ by the following fully connected neural networks: 
\begin{eqnarray}
  \bm{h}_t^{(1)}&=&\mathrm{relu}(\bm{W}_{t}^{(1)} \bm{h}_{t}+\bm{b}_{t}^{(1)}), \label{eq:y}\\
  \hat{\bm{y}_t}&=&\mathrm{softmax}(\bm{W}_{t}^{(2)} \bm{h}^{(1)}_{t}+\bm{b}_{t}^{(2)}),
%  \label{eq:y}
\end{eqnarray}
where $\bm{W}_t^{(1)}$ and $\bm{W}_t^{(2)}$ are weights and $\bm{b}_t^{(1)}$ and $\bm{b}_t^{(2)}$ are bias terms.

\subsection{Molecular Structure-based DDI Classification}

We represent drug pairs in molecular graph structures using two GCN methods: CNNs for fingerprints (NFP)~\cite{NIPS2015_5954} and Gated Graph Neural Networks (GGNN)~\cite{li2015gated}. They both convert a drug molecule graph $G$ into a fixed size vector $\bm{h}_g$ by aggregating the representation $\bm{h}^T_v$ of an atom node $v$ in $G$. 
We represent atoms as nodes and bonds as edges in the graph.

\textbf{NFP} first obtains the representation $\bm{h}^t_v$ by the following equations~\cite{NIPS2015_5954}.
\begin{eqnarray}
  \bm{m}^{t+1}_v&=&\bm{h}^{t}_{v}+\sum_{w \in{N(v)}}\bm{h}^t_w,\\
  \bm{h}^{t+1}_v&=&\sigma(\bm{H}_t^{deg(v)}\bm{m}_v^{t+1}),
\end{eqnarray}
where $\bm{h}^t_v$ is the representation of $v$ in the $t$-th step, $N(v)$ is the neighbors of $v$, 
and $\bm{H}_t^{deg(v)}$ is a weight parameter.
$\bm{h}^0_v$ is initialized by the \textit{atom features} of $v$. %, and $\bm{e}_{vw}$ is the representation of an edge between nodes $v$ and $w$ and initialized by the \textit{bond features}. The features and representations follow those proposed by \citet{rogers2010extended}.
$deg(v)$ is the degree of a node $v$ and $\sigma$ is a sigmoid function.
NFP then acquires the representation of the graph structure 
\begin{equation}
  \bm{h}_g=\sum_{v,t}\mathrm{softmax}(\bm{W}^t \bm{h}^t_v),
  \label{eq:gnfp}
\end{equation}
where $\bm{W}^t$ is a weight matrix.

\textbf{GGNN} first obtains the representation $\bm{h}^t_v$ by using Gated Recurrent Unit (GRU)-based recurrent neural networks~\citep{li2015gated} as follows:
\begin{eqnarray}
  \bm{m}^{t+1}_v&=&\sum_{w \in{N(v)}}\bm{A}_{e_{vw}}\bm{h}^t_w\\
  \bm{h}^{t+1}_v&=&\mathrm{GRU}([\bm{h}^t_v; \bm{m}_v^{t+1}]),
\end{eqnarray}
where $\bm{A}_{e_{vw}}$ is a weight for the \textit{bond type} of each edge $e_{vw}$.
GGNN then acquires the representation of the graph structure.
\begin{equation}
  %\bm{g}=\mathrm{tanh}(\sum_{v}\sigma(i([\bm{h}^T_v; \bm{h}^0_v]))\odot \mathrm{tanh}(j ([\bm{h}^T_v; \bm{h}^0_v])))
  \bm{h}_g=\sum_{v}\sigma(i([\bm{h}^T_v; \bm{h}^0_v]))\odot(j (\bm{h}^T_v)),
  \label{eq:gggnn}
\end{equation}
where $i$ and $j$ are linear layers and $\odot$ is the element-wise product.

We obtain the representation of a molecular pair by concatenating the molecular graph representations of drugs $g_1$ and $g_2$, i.e., $\bm{h}_{m}=[\bm{h}_{g_1}; \bm{h}_{g_2}]$. 

We get a prediction $\hat{\bm{y}}_m$ as follows:
\begin{eqnarray}
\bm{h}^{(1)}_{m} &=& \mathrm{relu}(\bm{W}^{(1)}_{m} \bm{h}_{m}+\bm{b}^{(1)}_{m}),\\
\hat{\bm{y}}_{m}&=&\mathrm{softmax}(\bm{W}^{(2)}_{m}\bm{h}^{(1)}_{m}+\bm{b}^{(2)}_{m}),
\end{eqnarray}
where $\bm{W}^{(1)}_{m}$ and $\bm{W}^{(2)}_{m}$ are weights and $\bm{b}^{(1)}_{m}$ and $\bm{b}^{(2)}_{m}$ are bias terms.

\subsection{DDI Extraction from Texts Using Molecular Structures}

We realize the simultaneous use of textual and molecular information by concatenating a text-based and molecule-based vectors: 
$\bm{h}_{all}=[\bm{h}_{t}; \bm{h}_m]$. 
We normalize molecule-based vectors. We then use $\bm{h}_{all}$ instead of $\bm{h}_{t}$ in Equation~\ref{eq:y}. % as shown in Figure~\ref{figure:model}. 

In training, we first train the molecular-based DDI classification model.
The molecular-based classification is performed by minimizing the loss function 
%\begin{eqnarray}
$L_{m} = -\sum{\bm{y}_{m} \log\hat{\bm{y}}_{m}}$.
%\end{eqnarray}
We then fix the parameters for GCNs and train text-based DDI extraction model by minimizing the loss function 
%\begin{eqnarray}
$L_{t} = -\sum{\bm{y}_{t} \log\hat{\bm{y}}_{t}}$.
%\end{eqnarray}

\section{Experimental Settings}
In this section, we explain the textual and molecular data and task settings and training settings.

\subsection{Text Corpus and Task Setting}

% \begin{table}[t!]
%   \centering\small
%   \begin{tabular}{lrr} \hline
%   & DRUG1 & DRUG2 \\\hline
%   Training data & 94.51 & 95.71 \\
%   Test data & 94.89 & 95.74 \\\hline
%   \end{tabular}
%   \caption{Matching rate (\%)}
%   \label{table:match}
% \end{table}

We followed the task setting of Task 9.2 in the DDIExtraction 2013 shared task~\cite{segura2013semeval,herrero2013ddi} for the evaluation. 
This data set is composed of documents annotated with drug mentions and their four types of interactions: \textsl{Mechanism}, \textsl{Effect}, \textsl{Advice} and \textsl{Int}. For the data statistics, please refer to the supplementary materials.
%As shown in this table, the number of pairs that have no interaction (negative pairs) is larger than that of pairs that have interactions (positive pairs).

%\todo{explicitly state that the evaluation of the DDIs is based on a multi-class classification task}
%Question 1: It is implied but not explicitly stated that the evaluation of the DDIs is based on a multi-class classification task, for the 4 distinct relation types. Is that the case?
%Yes, the evaluation is based on the multi-class classification task.
The task is a multi-class classification task, i.e., to classify a given pair of drugs into the four interaction types or no interaction. 
We evaluated the performance with micro-averaged precision (P), recall (R), and F-score (F) on all the interaction types. We used the official evaluation script provided by the task organizers.

As preprocessing, we split sentences into words using the GENIA tagger~\cite{geniatagger}. 
We replaced the drug mentions of the target pair with \textsl{DRUG1} and \textsl{DRUG2} according to their order of appearance.   
We also replaced other drug mentions with \textsl{DRUGOTHER}.
We did not employ negative instance filtering unlike other existing methods, e.g., \citet{liu2016drug}, since our focus is to evaluate the effect of the molecular information on texts.

We linked mentions in texts to DrugBank entries by string matching. We lowercased the mentions and the names in the entries and chose the entries with the most overlaps. 
As a result, 92.15\% and 93.09\% of drug mentions in train and test data set matched the DrugBank entries.

\subsection{Data and Task for Molecular Structures}
%\todo{clarify the binary classification task (it uses only the molecular information)}
%Question 2: Can you please clarify the binary classification task presented in Table 2? Does it use only the molecular information? (This is again implied, since "Text" isn't mentioned under Methods, but not clearly stated.)
%Yes, the binary classification task in Table 2 uses only the molecular information. The task is defined on drug entries in DrugBank and it has no corresponding text information.
We extracted 255,229 interacting (positive) pairs from DrugBank. 
We note that, unlike text-based interactions, DrugBank only contains the information of interacting pairs; 
there are no detailed labels and no information for non-interacting (negative) pairs.
We thus generated the same number of pseudo negative pairs by randomly pairing drugs and removing those in positive pairs. 
To avoid overestimation of the performance, we also deleted drug pairs mentioned in the test set of the text corpus.
We split positive and negative pairs into 4:1 for training and test data, and we evaluated the classification accuracy using only the molecular information. 

%\todo{add explanations on SMILES, RDKit, and input features}
%Reviewer 2
%Weakness argument 1: The notion of a graph representation for a pair of drug molecules is not explained and would be unknown and unintuitive to ACL readers.
%Thank you for the helpful comment. We will add more explanations on the representations from SMILES and RDKit and the features by Rogers and Hahn (2010).

%\todo{modify the figure}
\begin{figure}[t]
  \centering
  \includegraphics[width=1.\linewidth]{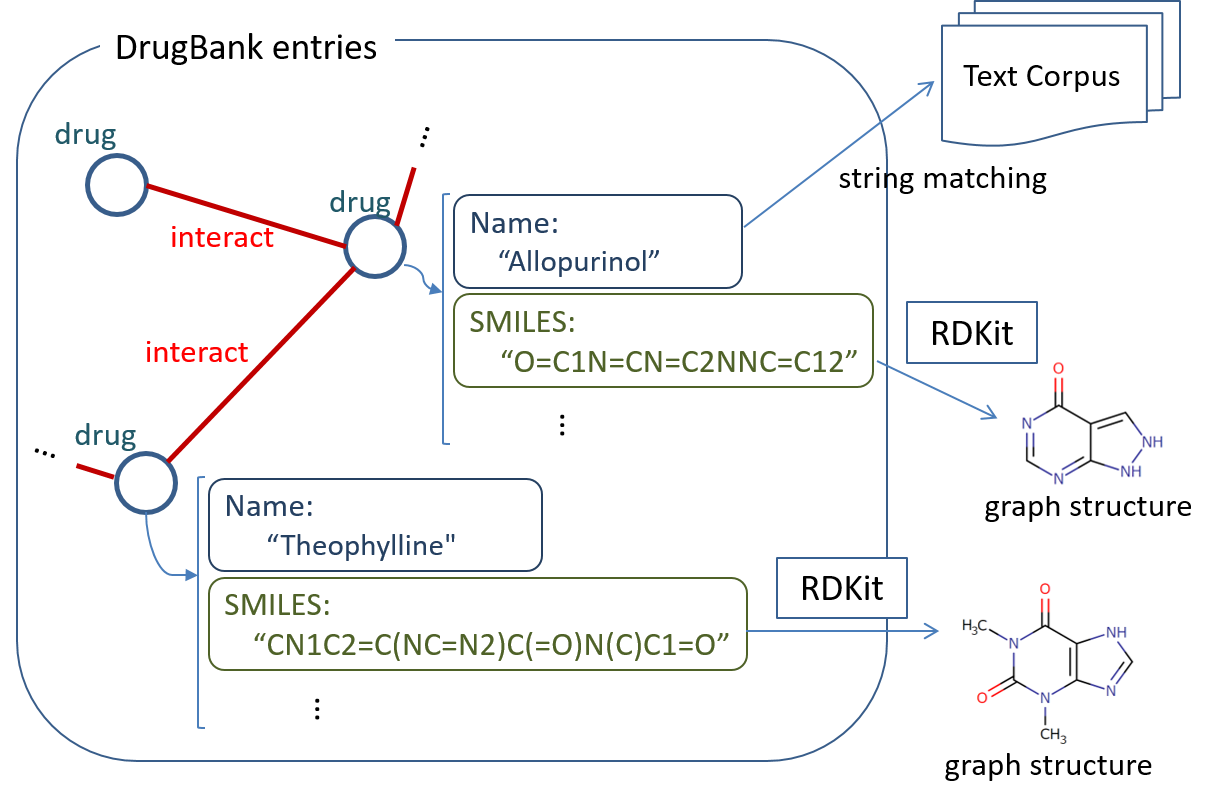}
  \caption{Associating DrugBank entries with texts and molecular graph structures}
  \label{figure:smiles}
\end{figure} 

To obtain the graph of a drug molecule, we took as input the SMILES~\cite{weininger1988smiles} string encoding of the molecule from DrugBank and then converted it into the graph using RDKit~\cite{landrum2016rdkit} as illustrated in Figure~\ref{figure:smiles}.
For the \textit{atom features}, we used randomly embedded vectors for each atoms (i.e., C, O, N, ...). We also used 4 \textit{bond types}: single, double, triple, or aromatic.

\subsection{Training Settings}

We employed mini-batch training using the Adam optimizer~\cite{kingma2014adam}. We used L2 regularization to avoid over-fitting. We tuned the bias term $\bm{b}^{(2)}_t$ for negative examples in the final softmax layer. 
%We randomly shuffled training data set and divide them into mini-batch samples in each epoch.
For the hyper-parameters, please refer to the supplementary materials. 

We employed pre-trained word embeddings trained by using the word2vec tool~\cite{mikolov2013distributed} on the 2014 MEDLINE/PubMed baseline distribution. The vocabulary size was 215,840.
The embedding of the drugs, i.e., \textsl{DRUG1} and \textsl{DRUG2} were initialized with the pre-trained embedding of the word \textsl{drug}.
The embeddings of training words that did not appear in the pre-trained embeddings were initialized with the average of all pre-trained word embeddings. Words that appeared only once in the training data were replaced with an \textsl{UNK} word during training, and the embedding of words in the test data set that did not appear in both training and pre-trained embeddings were set to the embedding of the \textsl{UNK} word. Word position embeddings are initialized with random values drawn from a uniform distribution.

We set the molecule-based vectors of unmatched entities to zero vectors. 

\section{Results}

%\todo{add discussions in the non-neural text-based relation extraction}
%Weakness argument 3: The authors do not acknowledge the body of work that leverages dependency graphs in relation extraction; although the older work involves non-neural models, it is relevant background. A few references are included below.
%[1] All-paths graph kernel for protein-protein interaction extraction with evaluation of cross-corpus learning https://doi.org/10.1186/1471-2105-9-S11-S2
%[2] A Dependency-Based Neural Network for Relation Classification https://arxiv.org/abs/1507.04646
%[3] Exploiting graph kernels for high performance biomedical relation extraction https://doi.org/10.1186/s13326-017-0168-3
%Thank you for the references. We will add discussions in the text-based relation extraction. 

\begin{table}[t!]
\centering
%\small
\begin{tabular}{p{3.5cm}lll} \hline
Methods & P & R & F (\%)\\\hline
%  \multicolumn{4}{c}{SVM-based models} \\\hline
%  \citet{chowdhury2013fbk} & 64.6 & 65.6 & 65.1 \\
%  \citet{kim2015extracting} & - & - & 67.0 \\
%  \citet{zheng2016graph} & - & - & 68.4 \\
%  \citet{raihani2017rich} & 73.60 & 70.07 & 71.79 \\\hline
%  \multicolumn{4}{c}{Neural network-based models} \\\hline
  \citet{liu2016drug} & 75.29 & 60.37 & 67.01\\
% \citet{sahu2017drug} & 67.77 & 66.80 & 67.28\\
  \citet{Zheng2017} & 75.9 & 68.7 & 71.5 \\
  \citet{lim2018drug} & 74.4 & 69.3 & 71.7\\\hline
  %\citet{lim2018drug} (ensemble) & 77.8 & 69.6 & 73.5\\\hline
%  Text-only  & 71.76 & 66.70 & 69.14\\
% + NFP & 67.60 & 73.75 & 70.54\\
% + GGNN & 68.89 & 73.75 & 71.24\\
% + NFP & 71.28 & 70.99 & 71.14\\
% + GGNN & 72.36 & 71.40 & 71.88\\
  Text-only  & 71.97 & 68.44 & 70.16\\
  + NFP & 72.62 & 71.81 & 72.21\\
  + GGNN & 73.31 & 71.81 & 72.55\\
  \hline
  \end{tabular}
  \caption{Evaluation on DDI extraction from texts}
  \label{table:comparison}
\end{table}

\begin{table}[t!]
\centering
%\small
\begin{tabular}{lllll} \hline
DDI Type & \textsl{Mech.} & \textsl{Effect} & \textsl{Adv.} & \textsl{Int} (\%)\\\hline
  Text-only & 69.52 & 69.27 & 79.81 & 48.18 \\
  + NFP & 72.70 & 72.44 & 79.56 & 46.98 \\
  + GGNN & 73.83 & 71.03 & 81.62 & 45.83 \\
  \hline
  \end{tabular}
  \caption{Performance on individual DDI types in F-scores}
  \label{table:type}
\end{table}

\begin{table}[t!]
  \centering
  \begin{tabular}{lr} \hline
  Methods & Accuracy (\%) \\\hline
  NFP &  94.19\\
  GGNN & 98.00 \\\hline
  \end{tabular}
  \caption{Accuracy of binary classification on DrugBank pairs}
   \label{table:bin}
\end{table}
\begin{table}[t!]
\centering
\begin{tabular}{lllll} \hline
Methods & P & R  & F & Acc. (\%)\\\hline
  NFP & 15.56 & 48.93 & 23.61 & 45.78\\
  GGNN & 15.11 & 57.10 & 23.90 & 37.72\\
  \hline
  \end{tabular}
  \caption{Classification of DDIs in texts by molecular structure-based DDI classification model}
  \label{table:detection}
\end{table}
%ASADA: 予測時にnegativeの項目に以下の数値を足す．
%no GCN: 0.004
%NFP : 0.210
%GCN : 0.185
%devel dataはofficial train dataをrelation typeごとに4:1に分割したもの

Table~\ref{table:comparison} shows the performance of DDI extraction models. We show the performance without negative instance filtering or ensemble for the fair comparison. We observe the increase of recall and F-score by using molecular information, which results in the state-of-the-art performance with GGNN. 

%\todo{do significance test}
%Weakness argument 2: The results are not reported with significance test.
%Thank you for pointing this out. We performed the approximate randomization test to compare our two methods with the text-only baseline model. We confirmed both models are significantly better than the baseline model (p < 0.05).  We include the results in the camera ready. 
Both GCNs improvements were statistically significant ($p<0.05$ for NFP and $p<0.005$ for GGNN) with randomized shuffled test. 
% シャッフル回数：10,000回
% NFP: p = 0.0168, GGNN: p = 0.0014

%\todo{show performance on individual DDI types}
%Weakness argument 2: Obviously this is a short paper so space is limited, but some error analysis of the performance across the 4 relation types is warranted; some types may benefit more from molecular information than others (e.g. Mechanism).
%We will definitely add the analysis of the performance on individual types.
Table~\ref{table:type} shows F-scores on individual DDI types. The molecular information improves F-scores especially on type \textsl{Mechanism} and \textsl{Effect}.

We also evaluated the accuracy of binary classification on DrugBank pairs by using only the molecular information in Table~\ref{table:bin}. 
The performance is high, although the accuracy is evaluated on automatically generated negative instances.

Finally, we applied the molecular-based DDI classification model trained on DrugBank to the DDIExtraction 2013 task data set. 
Since the DrugBank has no detailed labels, we mapped all four types of interactions to positive interactions and evaluated the classification performance. The results in Table~\ref{table:detection} show that GCNs produce higher recall than precision and the overall performance is low considering the high performance on DrugBank pairs. This might be because the interactions of drugs are not always mentioned in texts even if the drugs can interact with each other and because hedged DDI mentions are annotated as DDIs in the text data set. 
We also trained the DDI extraction model only with molecular information by replacing $\bm{h}_{all}$ with $\bm{h}_m$, but the F-scores were quite low ($<$ 5\%). These results show that we cannot predict textual relations only with molecular information.

\section{Related Work}

Various feature-based methods have been proposed during and after the DDIExtraction-2013 shared task~\cite{segura2013semeval}.
\newcite{kim2015extracting} proposed a two-phase SVM-based approach that employed a linear SVM with rich features that consist of word, word pair, dependency graph, parse tree, and noun phrase-based constrained coordination features. 
\newcite{zheng2016graph} proposed a context vector graph kernel to exploit various types of contexts. 
\newcite{raihani2017rich} also employed a two-phase SVM-based approach using non-linear kernels and they proposed five groups of features: word, drug, pair of drug, main verb and negative sentence features.
Our model does not use any features or kernels.  

Various neural DDI extraction models have been recently proposed using CNNs and Recurrent Neural Networks (RNNs).
\newcite{liu2016drug} built a CNN-based model based on word and position embeddings. 
%\newcite{sahu2017drug} proposed Joint AB-LSTM model that concatenates two RNN-based models: Bidirectional Long Short-Term Memory RNN (Bi-LSTM) and attentive pooling Bi-LSTM. 
\newcite{Zheng2017} proposed a Bidirectional Long Short-Term Memory RNN (Bi-LSTM)-based model with an input attention mechanism, which obtained  target drug-specific word representations before the Bi-LSTM. 
\newcite{lim2018drug} proposed Recursive neural network-based model with a subtree containment feature and an ensemble method. This model showed the state-of-the-art performance on the DDIExtraction 2013 shared task data set if systems do not use negative instance filtering. These approaches did not consider molecular information, and they can also be enhanced by the molecular information. 

\newcite{vilar2017detection} focused on detecting DDIs from different sources such as pharmacovigilance sources, scientific biomedical literature and social media. They did not use deep neural networks and they did not consider molecular information.

Learning representations of graphs are widely studied in several tasks such as knowledge base completion, drug discovery, and material science~\cite{wang2017knowledge,gilmer2017neural}. 
Several graph convolutional neural networks have been proposed such as NFP~\cite{NIPS2015_5954}, GGNN~\cite{li2015gated}, and Molecular Graph Convolutions~\cite{kearnes2016molecular}, but they have not been applied to DDI extraction.
%\newcite{gilmer2017neural} summarized existing models using their Message Passing Neural Network framework, and added variations to existing models. 

\section{Conclusions}

We proposed a novel neural method for DDI extraction using both textual and molecular information.
The results show that DDIs can be predicted with high accuracy from molecular structure information and that the molecular information can improve DDI extraction from texts by 2.39 percept points in F-score on the data set of the DDIExtraction 2013 shared task.

As future work, we would like to seek the way to model the textual and molecular representations jointly with alleviating the differences in labels. We will also investigate the use of other information in DrugBank.

\section*{Acknowledgments}

This work was supported by JSPS KAKENHI Grant Number 17K12741.

\bibliography{acl2018}
\bibliographystyle{acl_natbib}

\clearpage

\appendix
\section{Supplemental Material}
\subsection{Data Statistics}

The statistics of the data set is shown in Table~\ref{table:dataset}. This shows that the data is highly unbalanced and includes more negative pairs than positive pairs. 

\begin{table}[ht]
  \centering
  \begin{tabular}{llrr}\hline
  &DDI type & Train & Test\\\hline
  Positive & \textsl{Mechanism} & 1,319 & 302 \\
  & \textsl{Effect} & 1,687 & 360 \\
  & \textsl{Advice} & 826 & 221 \\
  & \textsl{Int} & 189 & 96 \\
  & Total & 4,021 & 979 \\
  Negative &  & 23,771 & 4,737 \\
  Total &  & 27,792 & 5,716 \\\hline
  \end{tabular}
  \caption{Statistics of the DDIExtraction 2013 shared task data set}
  \label{table:dataset}
\end{table}

\subsection{Hyper-parameter Settings}

We show hyper-parameters for training text-based model and molecular-based model in Tables \ref{table:param} and \ref{table:gcn_hyperparams}, respectively. 

\begin{table}[ht]
  \centering
  \begin{tabular}{lr}\hline
  Parameter & Value \\\hline
  Word embedding size & 200 \\
  Word position embedding size & 20 \\ 
  Convolution window size & [3, 5, 7] \\
  Convolution filter size & 100 \\
  Hidden layer size & 500 \\
  Initial learning rate & 0.001 \\
  Mini-batch size & 50 \\
  L2 regularization parameter & 0.0001 \\
  
  \hline
  \end{tabular}
  \caption{Hyperparameters for text-based model}
  \label{table:param}
\end{table}

\begin{table}[ht]
  \centering
  \begin{tabular}{lr} \hline
  Parameter & Value \\\hline
  Molecular vector size & 50 \\
  Number of steps & 4 \\
  Hidden layer size & 1,000 \\
  Initial learning rate & 0.001 \\
  Mini-batch size & 100 \\
  Hidden layer size of NFP & 50 \\
  GRU unit size of GGNN & 50 \\\hline
  \end{tabular}
  \caption{Hyperparameters for molecular-based model}
  \label{table:gcn_hyperparams}
\end{table}

\end{document}